\documentclass[conference,letterpaper,10pt]{ieeeconf}
\usepackage[left=1.91cm,right=1.91cm,top=1.91cm,bottom=1.91cm]{geometry}

\usepackage{graphicx}
\usepackage{amssymb}
\usepackage{amsmath}
\usepackage{cite}
\usepackage{comment}
\usepackage{color}
\usepackage{url}
\usepackage{alltt}
\usepackage{cleveref}
\usepackage{tikz}
\usepackage{pgfplots}
\usepackage{pgf-umlsd}
\usepackage{cprotect}
\usepackage{multirow}
\usepackage{algorithm}
\usepackage[noend]{algpseudocode}
\usepackage{bigfoot}
\usepackage{paralist}
\usepackage{tabularx}
\usepackage[whole]{bxcjkjatype}
\usepackage{ifthen}
\usepackage{threeparttable}
\usepackage[hang,small,bf]{caption}
\usepackage[subrefformat=parens]{subcaption}
\usepackage[super]{nth}
\usepackage{colortbl}
\usepackage{scalefnt}
\usepackage{threeparttable}
\usepackage{tabularx}
\usepackage{framed}

\pgfplotsset{compat=1.14}
\IEEEoverridecommandlockouts

\makeatletter
\newcommand{\figcaption}[1]{\def\@captype{figure}\caption{#1}}
\newcommand{\tblcaption}[1]{\def\@captype{table}\caption{#1}}
\makeatother

\title{\LARGE \bf
Four-Axis Adaptive Fingers Hand for Object Insertion: FAAF Hand
}

\author{Naoki Fukaya$^{1}$, Koki Yamane$^{2}$, Shimpei Masuda$^{1}$,\\
Avinash Ummadisingu$^{1}$, Shin-ichi Maeda$^{1}$, and Kuniyuki Takahashi$^{1}$
	\thanks{$^{1}$1 Preferred Networks, Inc.
		{\tt\small \{fukaya, masuda, ummavi, ichi, takahashi\}@preferred.jp}
              $^{2}$K. Yamane is with the University of Tsukuba. This work is done at Preferred Networks, Inc.
            {\tt\footnotesize 
                yamane.koki.td@alumni.tsukuba.ac.jp}
}
}

\begin{document}

   \onecolumn
   
    \begin{framed}
    © 2024 IEEE.  Personal use of this material is permitted. Permission from IEEE must be obtained for all other uses, in any current or future media, including reprinting/republishing this material for advertising or promotional purposes, creating new collective works, for resale or redistribution to servers or lists, or reuse of any copyrighted component of this work in other works.
    \end{framed}
    
    \twocolumn
    
    \newpage

\maketitle
\begin{abstract} 
Robots operating in the real world face significant but unavoidable issues in object localization that must be dealt with. A typical approach to address this is the addition of compliance mechanisms to hardware to absorb and compensate for some of these errors. However, for fine-grained manipulation tasks, the location and choice of appropriate compliance mechanisms are critical for success. 
For objects to be inserted in a target site on a flat surface, the object must first be successfully aligned with the opening of the slot, as well as correctly oriented along its central axis, before it can be inserted.
We developed the Four-Axis Adaptive Finger Hand ($\emph{FAAF hand}$) that is equipped with fingers that can passively adapt in four axes ($x, y, z, yaw$) enabling it to perform insertion tasks including lid fitting in the presence of significant localization errors. Furthermore, this adaptivity allows the use of simple control methods without requiring contact sensors or other devices.
Our results confirm the ability of the FAAF hand on challenging insertion tasks of square and triangle-shaped pegs (or prisms) and placing of container lids in the presence of position errors in all directions and rotational error along the object's central axis, using a simple control scheme.
\footnote{An accompanying video is available at the following link:\\ \url{https://www.youtube.com/watch?v=s3yf2MQ5Pag}}

\end{abstract}


\section{INTRODUCTION}
\label{sec:Introduction}
    When a robot operates in the real world, it needs to handle inevitable errors and uncertainty in the object pose due to numerous factors such as sensor noise, partial visibility, and prediction noise. This is particularly challenging in high-precision contact-rich tasks such as peg-in-hole and related insertion tasks that require consideration of the poses of both the peg and the hole.  
    
    Laboratory Automation is an active area of research in robotics due to the numerous open challenges it presents.  
    Typical tasks to be automated including inserting well-plates into dispensers and inserting lids onto well-pate and petri dishes are also highly demanding contact-rich insertion tasks with low clearances and margins of errors. 
    Failure in successful manipulation is also consequential due to the potential for breakage, spillage and contamination.
    There is also diversity in how different machines in a laboratory are to be loaded, limiting the application of mechanisms that only work in a vertical position (herein called the top-grasp position). 

    When considering the task of inserting an object into a target site fixed on a table, the most consequential relative positional errors occur along the 2 directions $(x, y)$ parallel to the table, the insertion height $(z)$, and the postural error along the central axis (yaw-axis). Significant errors along the $(x,y)$ directions ensure the object is not aligned with the slot, while errors in the $z$-axis direction may cause the object and target to collide or not even make contact. Even small errors along the yaw-axis may cause an inability of the object to be inserted into the hole without additional maneuvering to correct them.

\begin{figure}[t]
    \centering
    \includegraphics[width=0.95\columnwidth]{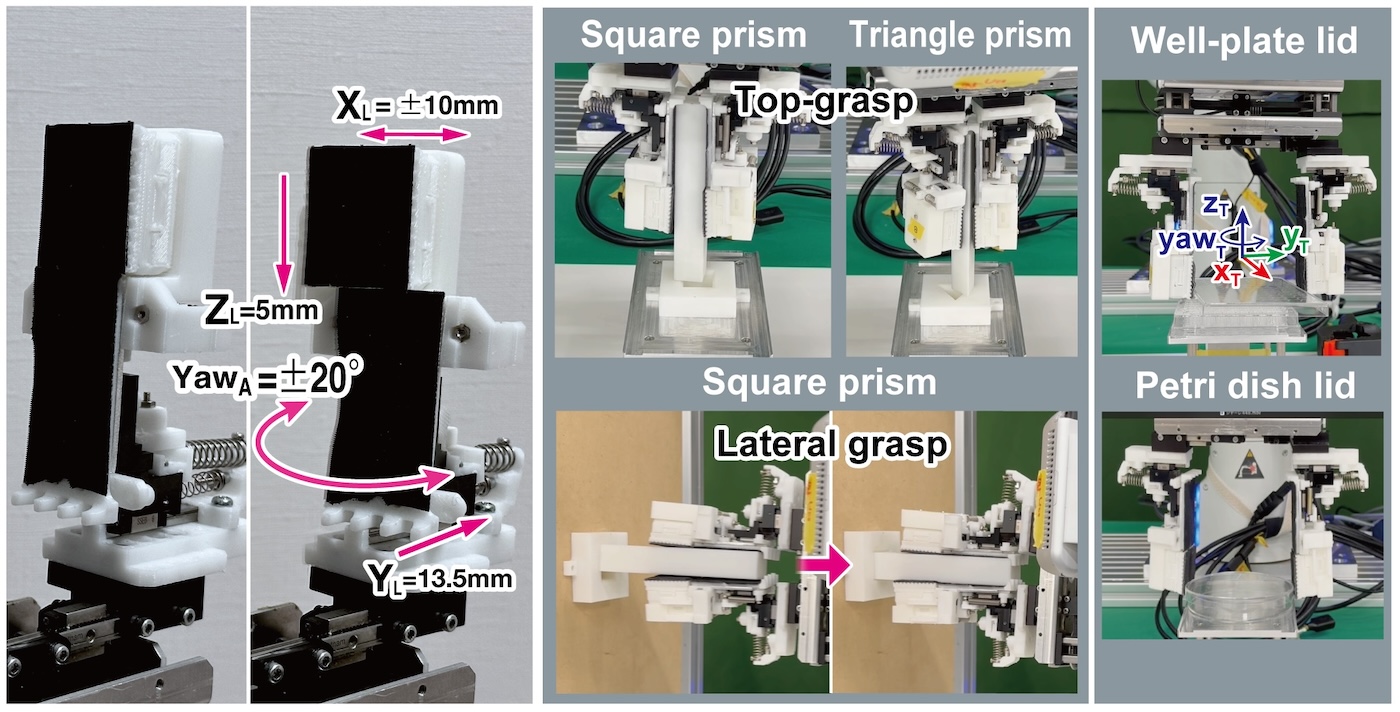}
    \caption{
    The FAAF hand has a four-axis passive mechanism in its fingers. 
    This enables simple controllers to perform challenging insertion tasks in the presence of pose errors.
    }
    \label{fig:topfigure}
\end{figure}

    One way to address problems of localization and misalignment is to provide a compliance function to the hardware.
    A popular method used is to have compliance at the wrist, which is between robot arm and gripper~\cite{Drigalski_6DOF_wrist_OSX, Zhang_6DOF_spral_PEG_insert}.
    However, the high-mass wrist may become unstable due to gravitational influences in postures other than the vertical grasp position. 
    On the other hand, having compliance at the finger makes them lighter and less affected by gravity. 
    
    One approach is to use a finger fully made of soft, deformable material~\cite{Subramaniam_soft_gripper2020, kuppuswamySoftbubbleGrippersRobust2020}.
    These methods are challenging and cannot achieve object positioning accuracy, making high-precision contact-rich tasks difficult.
    Alternatively, methods of solid material fingers with compliance mechanisms built into finger joints have been proposed to achieve insertion tasks. 
    \cite{Morgan_vision_compliant_assembly} verified that increased compliance in the hand and robotic arm boosts the success rate of peg-in-hole tasks. This approach requires in-hand manipulation where a vision-driven servoing framework controls the postures of the fingers. Nevertheless, even with multi-finger in-hand manipulation, inserting triangles or rectangles with yaw axis errors remains a challenge.
     
    By using hardware adaptive features, it can be possible to avoid using complex controls without relying on vision and sensors~\cite{Fukaya_F2_hand}.
    \cite{Fukaya_F2_hand} confirmed that a compliant hand driven by a single motor is capable of inserting cylindrical pegs with a simple control system.
    However, this hand does not support in-hand manipulation, making it challenging to insert triangles, squares, and other shapes with yaw axis errors.
    This suggests that, in addition to the $xyz$ positional error, addressing posture error around the yaw axis is crucial for successfully managing the insertion movements of objects other than cylindrical ones.
    
    To summarize the above issues, the requirements for a hand to achieve contact-rich insertion tasks which require precise manipulation are as follows:
    \begin{enumerate}
        \item Compensate for positional errors in all 3 directions $(x, y, z)$ and one postural (yaw-axis).
        \item Maintains sufficient stiffness with compliance mechanisms built into finger's joints to perform precision peg-in-hole and lid-placement tasks with small clearance.
        \item Is lightweight enough to allow operation in postures other than the top-grasp position.
         \item Realization of insertion task by simple control using the sensor-less robot fingers.
    \end{enumerate}
    
    To realize these requirements, we developed the sensor-less \textbf{F}our-\textbf{A}xis \textbf{A}daptive \textbf{F}inger \textbf{Hand} （\emph{FAAF Hand}), which has passive mechanisms for hardware compliance in the $xyz$-axis + yaw-axis direction in the finger component (Fig.~\ref{fig:topfigure}). 
    We conducted experiments to verify whether these mechanisms aid in achieving our tasks (Peg-insertion of square prisms and triangular prisms, and insertion of well-plate's lid and perti dish's lid), their tolerance to noise and errors, as well as study which one of these adaptations is crucial in each case.

\section{RELATED WORK}
\label{sec: related_work}

Several studies have been conducted to provide hardware structures to the robot to address situations where localization errors exist. One approach is to absorb misalignment caused by recognition errors by providing compliance to the wrist and finger.

\subsection{Wrist Compliance Mechanisms }
A necessary task for which precise positioning of the grasped object is important is the peg insertion task. 
One approach is based on the Remote Center
Compliance (RCC) and is able to compensate for a misalignment of 2~mm during insertion by supporting the wrist with an elastic body~\cite{Joo_RCC, Choi_VRCC, Lee_VRCC}.
For major misalignments, other studies proposed using softer springs at the wrist. \cite{Royo_Peg_in_hole_gelsight_OSX} demonstrated that it is feasible to insert a peg into a hole that has shifted by 10~mm.
\cite{Zhang_6DOF_spral_PEG_insert} showed that the hand unit can move passively in 6 axes to insert a common cylindrical peg with a misalignment of 11~mm and other shapes such as a square or triangular prism, with a misalignment of 4~mm.
For triangular prisms in particular, insertion was achieved with an additional 10-degree rotation around the long axis.
Since these methods perform the insertion operation with the hand unit suspended, if the fingertip is pointed in a non-vertical direction, the hand unit may hang down due to its own weight, making it difficult to follow the target trajectory.
Therefore, when the insertion motion is performed in a posture other than the top grasp, the mass of the passively moving part should be as light as possible.

\subsection{Finger Joint Compliance Mechanisms}
There is an abundance of research on conferring compliance to the finger to alleviate the need for precise control of finger joint movements. One approach is to build part or all of the finger with elastic materials. \cite{Takahashi_ring_insert} provides a self-aligning mechanism to the gripper that combines a shaft and a deformable elastic ring.  
\cite{Hartish_3dprint_connector_insert_hand} proposed a method to insert connectors using soft fingers created by a 3D printer. Despite the need to design and fabricate a finger for each connector, this method has the advantage of simplicity since the entire finger is constructed with a 3D printer.
\cite{Zao_gelsight_finger_hand} developed a hand made with a flexible material that serves as a tactile sensor for the whole finger, enabling elastic deformation of the fingertip in numerous degrees of freedom, including the yaw direction. However, once the finger is pliable, controlling this elastic deformation becomes challenging. Therefore, instead of forming the finger as a soft body, studies have explored methods that bestow compliance on the motion mechanism side.
\cite{DefGear_qiujie_gripper_2021} proposed utilizing the fingertip's adaptability to grasp a misaligned object. However, with the object's posture changing post-grasp, it is difficult to accurately place the object during placing.
The F2 hand, equipped with an adaptive mechanism assembled with an aluminum link mechanism, can grasp an object off-center and insert a peg in the offset position~\cite{Fukaya_F2_hand}. To guide a peg into an off-center hole, the fingers must offer resistance during insertion- made possible owing to their rigid construction. However, with rotating fingers, the fingertip angle tilts during off-center grasping, complicating the lifting of plate-like objects like well-plates horizontally. Also, this finger has no yaw-axis adaptation function.

\section{DESIGN OF FAAF HAND}
\label{sec:hand_design}

\begin{figure}[t]
    \centering
    \includegraphics[width=0.9\columnwidth]{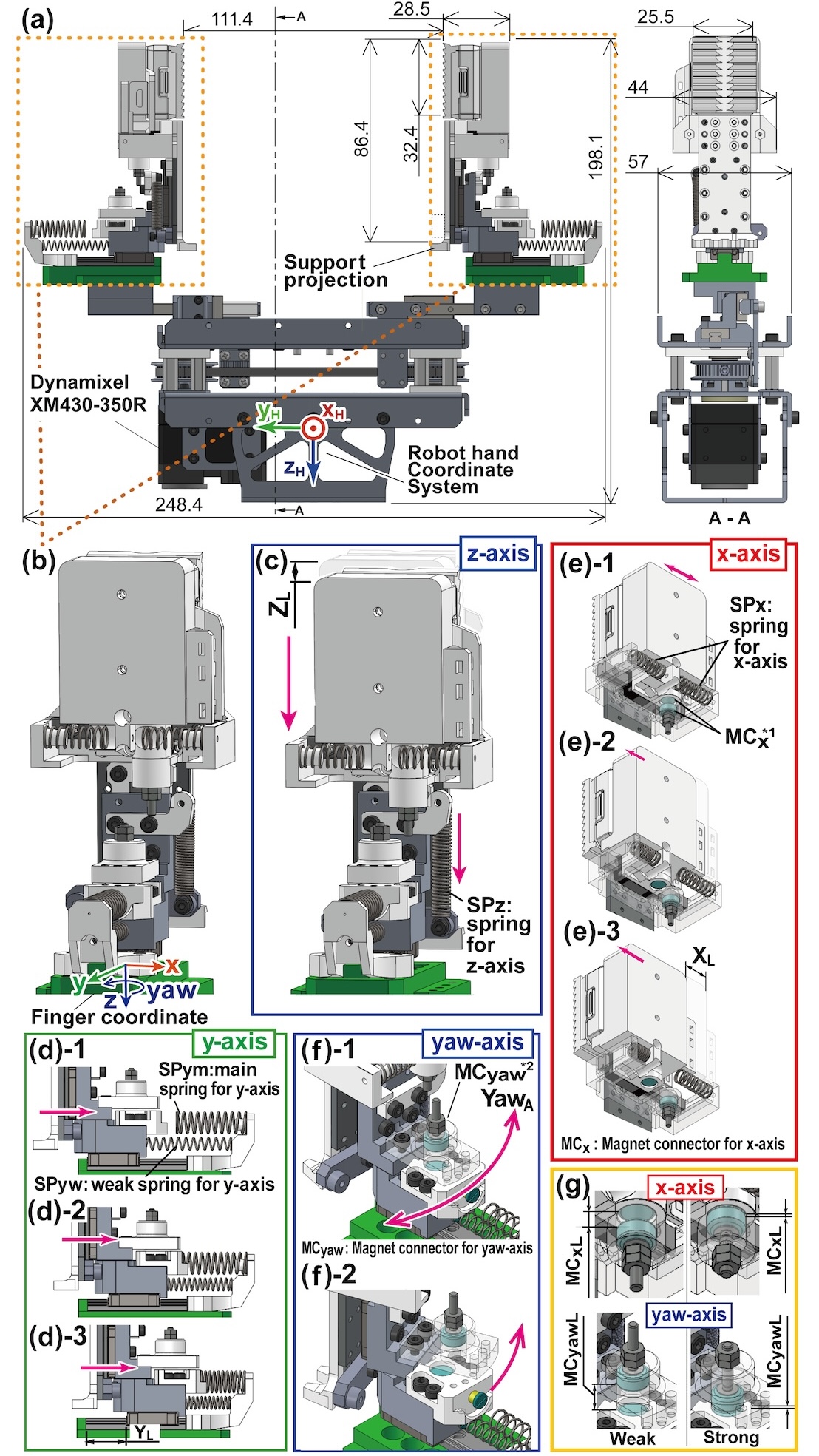}
    \caption{
        (a) Hand structure and dimensions. (b) Finger with 4-axis adaptive mechanism. (c)-(f) Structures for adaptive movement in the $z$-, $y$-, $x$-, and yaw-axis directions. (g) Magnet connector (MC) can change the reaction force as required by changing the distance between the two magnets.
    }
    \label{fig:adaptive_motion}
\end{figure}

Fig.~\ref{fig:adaptive_motion}~(a) presents a comprehensive view of the robotic hand. The main structure incorporates a commonly-used parallel gripper configuration, equipped with a two-stage linear slider, which governs the motion of opening and closing of the fingers. The XM430-350R motor from Dynamixel is utilized for this purpose. The motor's position control dictates the degree of openness or closedness of the fingers. Note that the robot hand's adaptive mechanisms are localized strictly to its finger section. The coordinate axes of the hand is defined by ($x_H, y_H, z_H, yaw_H$).

For the finger component, we base our design on a prior version of our finger~\cite{Sameer_well_plate_set} that allows adaptation to the $y$-axis and $z$-axis. To allow further adaptability in the $x$-axis and yaw-axis directions, additional adaptive mechanisms were incorporated for these two axes.
The internal end of the finger has a support projection for the grasped object (Fig.~\ref{fig:adaptive_motion}~(a)). This projection also moves with the adaptive movement in the $z$-axis direction.
These four axes ($x,y,z, yaw$) follow the finger coordinate system (Fig.~\ref{fig:adaptive_motion}~(b)).
During this experiment, the finger surfaces were covered with a TB631 (3M) gripping tape.
These fingertips are equipped with a double layer of this tape.


\begin{figure}[t]
    \centering
    \includegraphics[width=0.9\columnwidth]{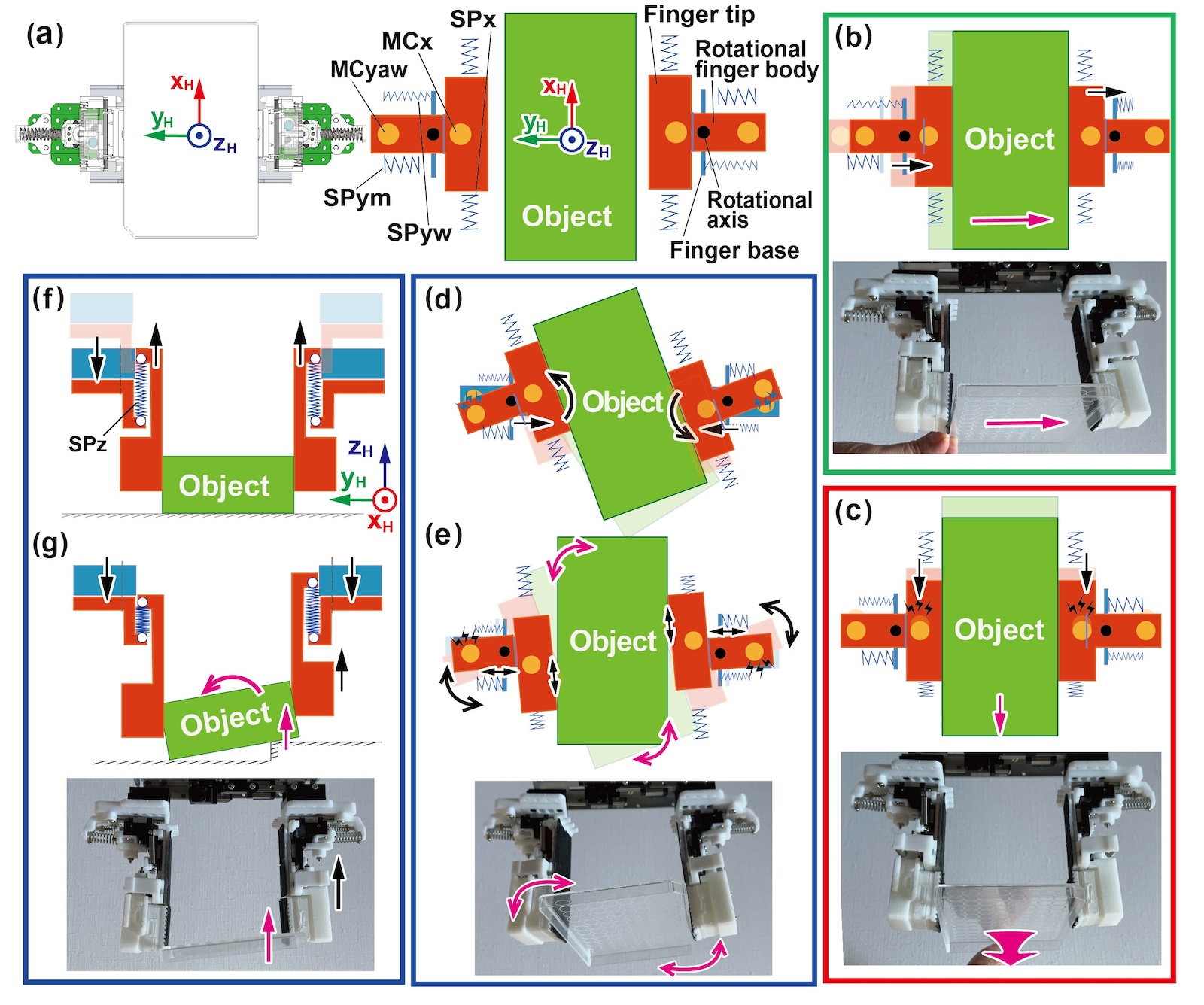}
    \caption{
    The finger's adaptive mechanisms in action. (a) Initial state. (b)-(c) Motion when the grasped object moves in the $y$- or $x$-axis direction. (d) Grasping object operation to arbitrarily change the posture of an object in response to an external force. (e) Motion when an object rotated along its yaw-axis, object is grasped and subjected to external force and its posture changes. (f)-(g) The left and right sides move independently and can adjust to the undulations of the ground. 
    }
    \label{fig:adaptive_grasp_work}
\end{figure}

\subsection{Adaptive Mechanism of the $y$-axis and $z$-axis}
\label{sec:y_z_adaptive_mech}
Fig.~\ref{fig:adaptive_motion}~(c) shows the adaptive motion in the $z$-axis direction.
The amount of slide in the $z$-axis direction is 5~mm ($Z_L$) and the spring constant is 0.15~N/mm.

For the $y$-axis, the slide volume was increased by 2.7 times to 13.5~mm ($Y_L$) because the structure in~\cite{Sameer_well_plate_set} has a limited sufficient adaptable range (Fig.~\ref{fig:adaptive_motion}~(d)-1).
This structure is designed to displace the object being grasped, in the $y$-axis direction. For the grasped object to swing equitably to the left and right in the $y$-axis direction, the fingers must be positioned at the center of the slide width (Fig.~\ref{fig:adaptive_motion}~(d)-2). At this point, the main spring must exert adequate reaction force on the fingers to grasp the object. However, if the reaction force is generated from the starting stage, the reaction force occurring at the midpoint will increase. If we attempted to decrease the spring constant, the rate of increase in reaction force at the midpoint would remain low. However, this would mean it becomes challenging to attain sufficient reaction force at the end of the slide. Therefore, we have combined two types of springs: a main spring (spring constant: 0.67~N/mm) to provide ample reaction force from the midpoint to the end of the slide and a weaker follow-up spring (spring constant: 0.15~N/mm) from the start point up to the midpoint to return the finger to its initial position.
Note that if a strong gripping force is employed, it nullifies the adaptive function in the $y$-axis direction, thus allowing the object to be firmly grasped in a manner akin to using a conventional gripper (Fig.~\ref{fig:adaptive_motion}~(d)-3).


\subsection{Adaptive Mechanism of the $x$-axis and yaw axis}
\label{sec:x_yaw_adaptive_mech}

The $x$-axis is designed for a 10~mm slide to the left and right ($X_L$)(Fig.~\ref{fig:adaptive_motion}~(e)).
Regarding the reaction force at the initial and nearby positions of the fingers, we developed a structure (hereafter addressed as Magnet Connector (\emph{MC}) incorporating two Neodymium magnets (d=6~mm, t=3~mm, magnetic flux density 280~mT) (Fig.~\ref{fig:adaptive_motion}~(e)-1). 
The reaction force, triggered by the magnetic force of the magnets, operates within a range of up to $ \pm$4~mm when a load is applied. 
When the magnets are placed too far apart, the reaction force ceases to exist  (Fig.~\ref{fig:adaptive_motion}~(e)-2). To account for this, a push spring is introduced to generate reaction force beyond this point. This spring possesses a constant of 0.78$\mathrm{\,N/mm}$.
The yaw-axis also utilizes the MC mechanism to simplify its structure and control the reaction force. It has a rotation angle of $\pm20$ degrees ($Yaw_A$).
It allows for contact with lightweight or easily deformable objects without the need for the object to be moved or deformed - a benefit most noticeable when the reaction force is kept to a low level (Fig.~\ref{fig:adaptive_motion}~(f)).
The distance between the two magnets of MC can be adjusted ($MC_{xL}$=0.5-3~mm, $MC_{yawL}$=1-5~mm so that the repulsive force can be easily changed according to the purpose or application  (Fig.~\ref{fig:adaptive_motion}~(g)).
These two axes passively adjust the fingertip's posture when the fingertip's surface and the object are not parallel. This increases the contact area and enhances the grasp stability (Fig.~\ref{fig:adaptive_grasp_work}~(b)~\&~(c)). Moreover, when an external force is applied to the grasped object, they enable in-hand manipulation. This allows the object's posture to change according to the external force without releasing the object (Fig.~\ref{fig:adaptive_grasp_work}~(d)).

\begin{figure}[t]
    \centering
    \includegraphics[width=0.9\columnwidth]{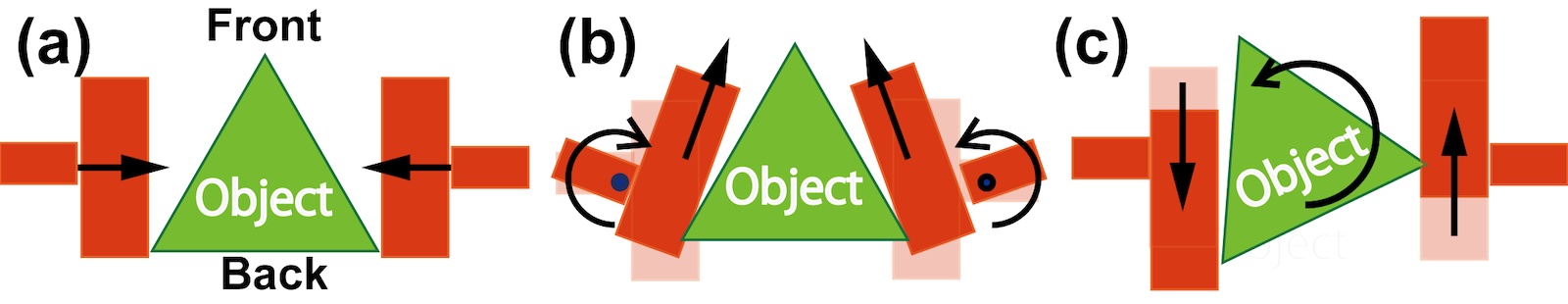}
    \caption{
    (a) Grasping a triangle. (b) The fingertip can rotate to increase the contact area by yaw and $x$-axis adaptive function. (c) The adaptive function of the $x$-axis alone can rotate the grasped object.
    }
    \label{fig:x_Yaw_function}
\end{figure}

\subsection{Reaction Forces along Each Axis}

\begin{figure}[t]
    \centering
    \includegraphics[width=0.9\columnwidth]{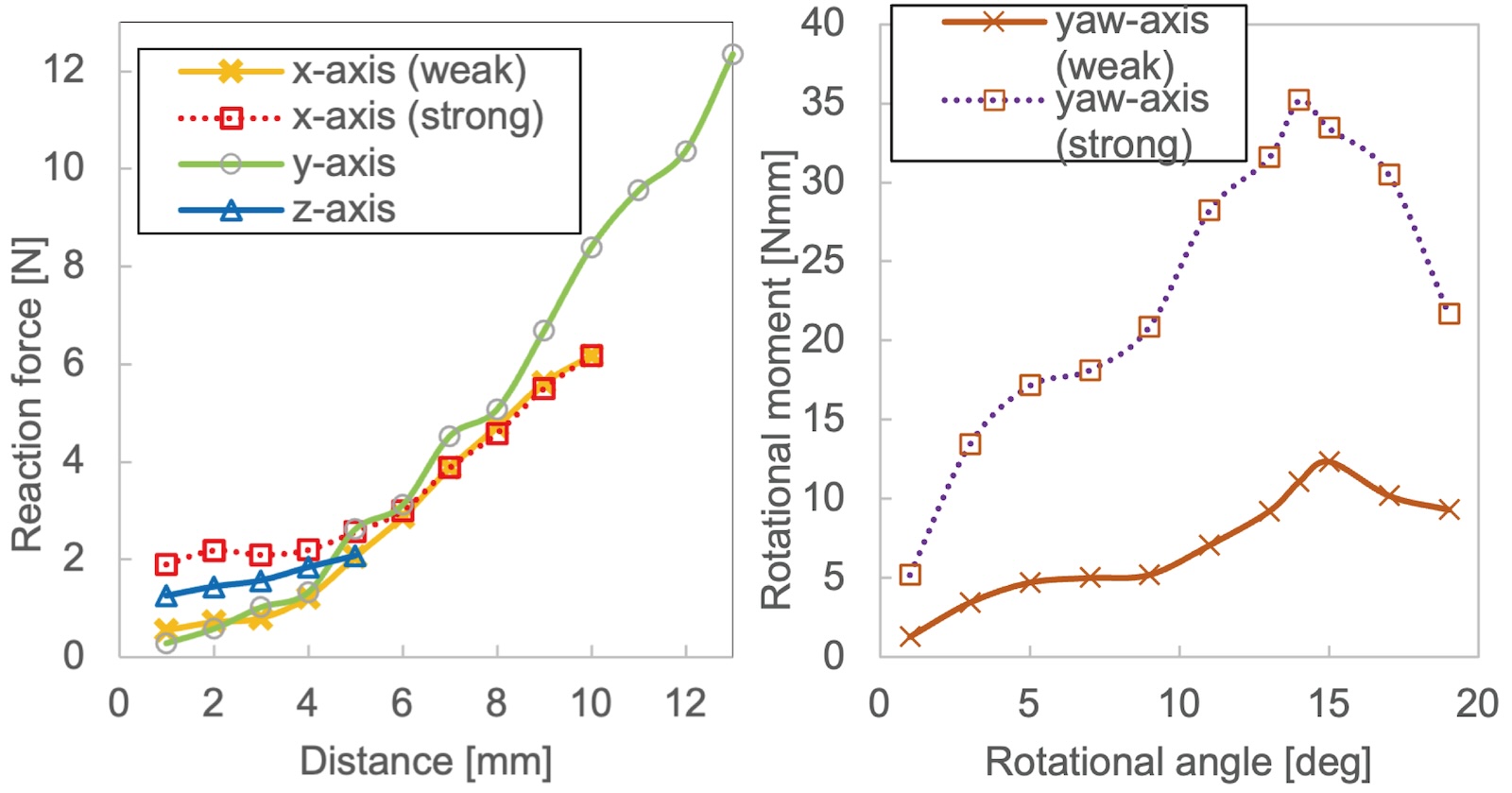}
    \caption{
     Reaction force for each axis.
    }
    \label{fig:reaction_force}
\end{figure}

Every axis in the finger coordinate experiences a reaction force affected by either a spring or the magnet's magnetic force. The maximum reaction force acting on the $y$-axis and $x$-axis has been set high as these are necessitated to adapt during the process of object grasping. Conversely, the maximum levels of reaction force on the $z$-axis and yaw-axis, which primarily function to modify the finger's posture upon making contact with a physical object or the environment, are purposefully set low. Fig.~\ref{fig:reaction_force} shows the reaction force present on each axis as measured by a digital force gauge (DFG-2K, Shinpo). 
MC on the $x$-axis and yaw-axis is basically used with the two magnets at a farther distance from each other and with the reaction force weakened (weak). For reference, the state in which the magnets are closest to each other and the reaction force is stronger (strong) is also shown.
The $x$-axis experiences a reaction force generated by the centrally-located magnet. As the magnet moves further from this center, its magnetic force weakens, causing a slight dip in the reaction force. However, upon reaching a distance of 5~mm, contact with the coil spring instigates an increase in the reaction force.
In terms of the $y$-axis, the weak holding spring functions solely up until the 5~mm mark. Beyond this, the main spring comes into work, resulting in an increased rate of reaction force beyond this position.
Since the distance between the magnets does not change as much as in the $x$-axis, the peak of the magnetic force exists around 14 degrees.

\section{EXPERIMENTAL SETUP}
\label{sec: experimental_conditions}

\subsection{Robotic System}
\begin{figure}[t]
    \centering
    \includegraphics[width=0.9\columnwidth]{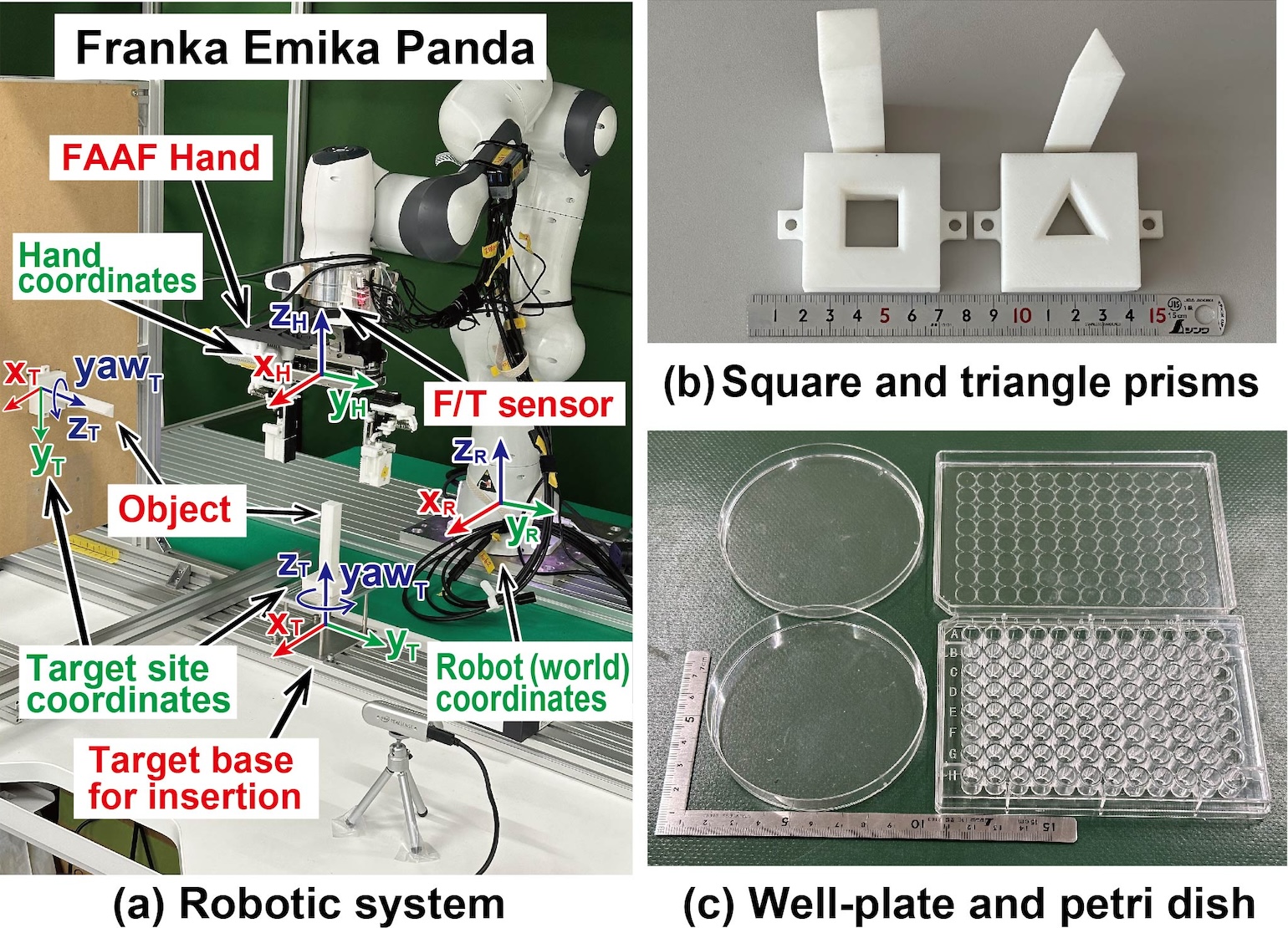}
    \caption{
    Overview of the experimental environment and objects used in the experiments.
    }
    \label{fig:experimental_setup}
\end{figure}

The Franka Emika Panda Arm 7-DoF robotic arm was used for the experiments. The coordinate system conforms to the robot coordinate system ($x_R, y_R, z_R$) (Fig.~\ref{fig:experimental_setup}). The control frequency is 1~kHz. A Leptrino force/torque (F/T) sensor (FFS055YA501U6) was attached between the robot arm and the FAAF hand. The F/T sensor detects the force applied to the hand: $F_{xyz}= \pm$500~N for force, $T_{xyz}= \pm$4~Nm for torque, with a resolution of $\pm$1/2000, and a sampling rate of 200~Hz.
The control PC consists of 32 GB RAM, Intel Core i7-7700K CPU, and Ubuntu 20.04.6 LTS with ROS Noetic installed.
We create a \emph{target base} area where objects to be experimented with are inserted. For well-plate insertion experiments, the top surface of the holder. The top surface of the holder is surrounded by a raised groove of height 1.5~mm. The dimensions of the inside edges are 128.2~mm and 86.2~mm~\cite{Sameer_well_plate_set}. Each coordinate axis of the experimental object conforms to the target site coordinate system ($x_T$, $y_T$, $z_T$, $yaw_T$).

\subsection{Objects Used in the Insertion Experiments}
\label{sec: setup_for_insertion}

\begin{figure}[t]
    \centering
    \includegraphics[width=0.9\columnwidth]{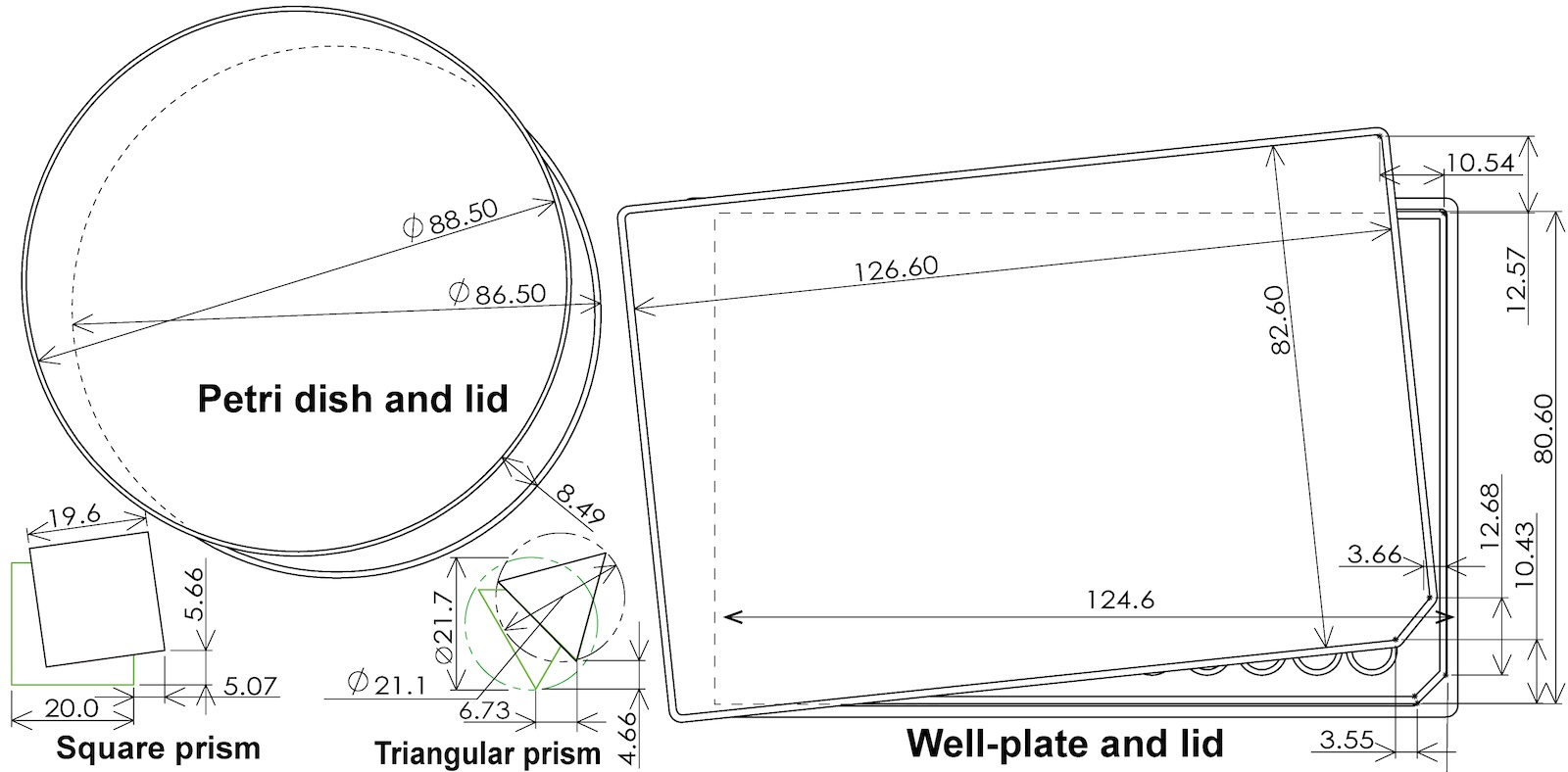}
    \caption{
    Peg and Lid insertion task with an example of an initial offset position used during experiments.
    }
    \label{fig:well_plate_2_Peg}
\end{figure}

For insertion experiments involving position errors of the object's yaw-axis in the target site coordinate system ($yaw_T$) in Fig.~\ref{fig:experimental_setup}, we create two types of non-cylindrical pegs (square and triangular prism).
Both pegs and insertion bases were fabricated from ABS material using a 3D printer referring to \cite{Zhang_6DOF_spral_PEG_insert}.
The insertion bases are designed to be only slightly larger to highlight the impact of posture error.
The square prism has dimensions of 19.6~mm on each side and a length of 120~mm. The corresponding insertion base has a side length of 20.0~mm, with chamfers of 2~mm. The triangular prism has a shape with the radius of the circumscribing circle measuring 21.1~mm and a length of 120~mm. The corresponding insertion base has a radius of 21.7~mm, with chamfers of 2.5~mm as shown in Fig.~\ref{fig:well_plate_2_Peg}. The depth of the insertion bases for both the square prism and the triangular prism is 15~mm.
The pegs were left unchamfered.

Well-plates (GDMP-96F, ASONE), their lids (GDMP-CV/2, ASONE), and petri dishes were used for the lid insertion experiments (Fig.~\ref{fig:experimental_setup}).
The inner dimensions of the well-plate lid are 124.6~mm $\times$ 82.6~mm. The depth of the lid is 8.2~mm, and the clearance between the well-plate and the lid is 2~mm. 
The well-plate measures 127.3~mm in length, 85~mm in width, and 14.4~mm in height. 
One corner is chamfered by 7~mm to prevent misorientation when inserting. 
The petri dish lid is 13.1~mm in depth, with an internal diameter of 88.5~mm, a thickness of 0.8~mm, and a depth of 11.9~mm. The clearance of the lid to the petri dish is 2~mm.

\subsection{Insertion Limit Angle for Yaw-rotated Objects}
\label{sec:exp_set_insert_angle_Yaw}
To ascertain the maximum insertion angle for an object rotated along its $yaw_T$-axis, variants of the peg and lid insertion tasks were created using two types of non-cylindrical pegs and well plate lids (\cref{sec: setup_for_insertion}).
These objects were picked up vertically, rotated by an arbitrary degree along their yaw-axis, and inserted directly into the receiving section. The maximum angle in the $yaw_T$-axis direction at which the object could be successfully inserted without moving the object laterally is thereby identified.
The $yaw_T$ was incremented by 1 degree, starting from 0 degree. The angle at which the objects could be successfully inserted five consecutive times was determined.

\subsection{Peg Insertion}
\subsubsection{Peg Insertion (Vertical)}
\label{sec:exp_set_peg_insert}

To validate the effectiveness of the proposed method, insertion experiments were conducted using two types of pegs. Upon being grasped by the FAAF hand, the peg was lifted vertically and rotated an arbitrary degree around the $xy$-axis ($x_T, y_T$) and yaw-axis ($yaw_T$) before the insertion operation was initiated. This degree of displacement is hereafter referred to as the offset. 
To insert the peg from this raised position, the robot moved the peg in a spiral path, referring to \cite{Zhang_6DOF_spral_PEG_insert}. 
Note that no tilt angle was given to the pegs so that it is not affected by changes in peg length or grasping position.
The spiral path used was the Archimedes' spiral. Assuming that the coordinates of the robot hand when attempting to grasp an object are (0, 0, $z_T$) in the target site coordinate system, the parametric variable display of this spiral path is as follows:
\begin{eqnarray}
    R_{sp} = (R_{start} – R_{end})((Dr – t) / Dr) + R_{end} \\
    x_{sp}= R_{sp} \cos ((2 \pi R_{num} t) / Dr) + x_{offset} \\
    y_{sp} = R_{sp} \sin ((2 \pi R_{num} t) / Dr) + y_{offset}
\end{eqnarray}
Here, $t$ is unit time, $D_r$ is time duration from start to end, $\mathrm{R_{start}}$ is the start radius of spiral, $\mathrm{R_{end}}$ is the end radius of spiral, $\mathrm{R_{num}}$ is the number of circumferences, $\mathrm{x_{offset}}$ and $\mathrm{y_{offset}}$ are offset amount in $x_T $ and $y_T$ direction respectively. $z_T$ is the arbitrary height at which each object is grasped.

\begin{figure}[t]
    \centering
    \includegraphics[width=0.85\columnwidth]{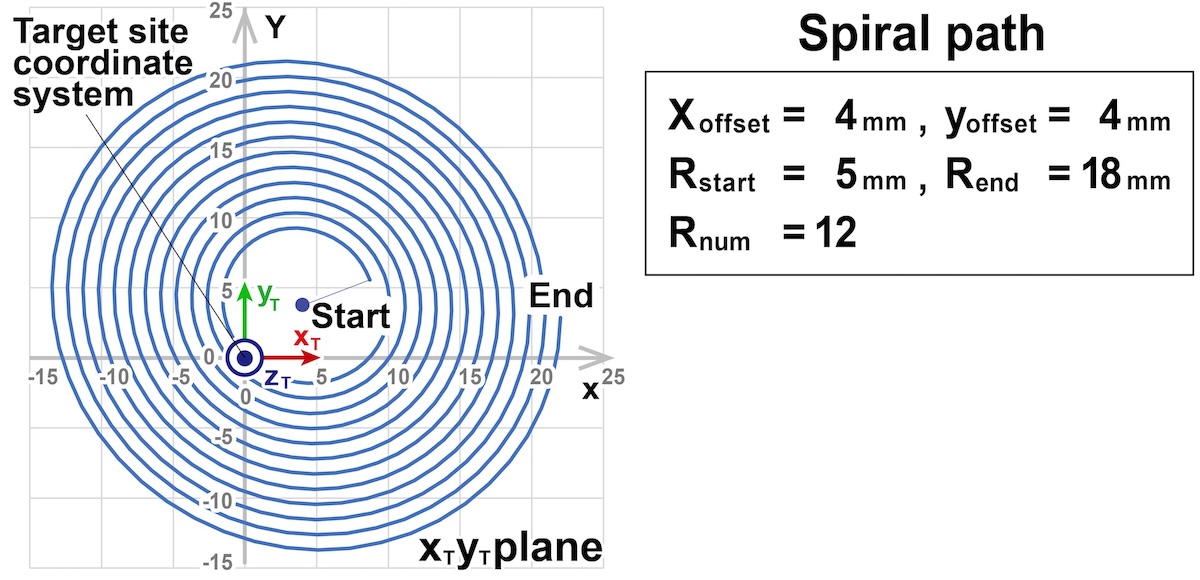}
    \caption{
   An example of a spiral path attempting to place a square prism from an offset position.
    }
    \label{fig:exp_spral_curve}
\end{figure}

As the peg passes over the insertion site along this path, the adaptive mechanism automatically allows the peg to change position in the hand so it can be inserted.
An example of the spiral path used in the experiment is shown in Fig.~\ref{fig:exp_spral_curve}.
The spiral path rotates for a fixed number of revolutions while increasing the radius of rotation from the start radius to the end radius.
Instead, it was held vertically and ushered into contact with the surface while tracing a spiral path after moving to offset position. 
The contact force between the peg and the contact surface was maintained at a constant value, as determined by the F/T sensor.
Eight combinations were set for the offsets, with positive and negative values in the $x_T$-axis, $y_T$-axis direction, and around the $yaw_T$ axis, respectively.
Success was determined when all four sides of the insert's bottom descended below the chamfer by the completion of the spiral path motion.

In order to verify the functioning of the compliance mechanisms, comparative experiments were conducted with the compliance mechanisms held fixed individually for each dimension. The deactivation of each compliance mechanism was achieved as follows. The $z$-axis and $y$-axis were disabled by inserting parts that inhibit their operation, the $x$-axis was replaced by parts without compliance mechanisms, and the yaw-axis was held fixed with screws to disable each adaptive mechanism.

\subsubsection{Peg Insertion (Lateral)}
\label{sec:exp_set_lateral_insert}

To validate the benefits of incorporating the adaptive mechanism into the lightweight finger section, as opposed to the wrist, we conducted an experiment that involved inserting the peg horizontally. To eliminate any effects of posture modifications, we kept the conditions for the pegs and the rotation radius of the spiral path the same as previously described in \cref{sec:exp_set_peg_insert}.

\subsection{Lid Insertion}
\label{sec:exp_set_lid_insert}

The well-plate and petri dish (Fig.~\ref{fig:experimental_setup}) were set in a holder with the lid in place. 
The lid was grasped by the FAAF hand, lifted, offset added, and then attempted to insert with a spiral path as in \cref{sec:exp_set_peg_insert}. The lid was pressed by a constant value force modulated by the F/T sensor in the same method of peg-insertion. 
Success was determined when the bottom edge of the lid had descended below the top of the target site object by the completion of the spiral path motion.
To confirm the effectiveness of the adaptive functions, we also checked the success rate by locking the adaptive function of each of the four axes of the finger.

\section{EXPERIMENTAL RESULT}
\label{sec: experimental_result}

\subsection{Insertion Limit Angle for Yaw-rotated Objects}
\label{sec:result_insert_angle_Yaw}

The experimental results indicated that the maximum offset angle in the $yaw_T$-axis direction, which permits insertion, was 5 degrees for the square prism, 9 degrees for the triangular prism, and 3 degrees for the well-plate lid. The larger tolerance for the triangular prism can be attributed to the difference in hole chamfer size in addition to the variance in geometry. The well-plate lids, due to their elongated length, were more sensitive to angle variation, resulting in a smaller tolerance.

\subsection{Peg Insertion}
\subsubsection{Insertion Experiment of Square Prism}
\label{sec:result_square_prism}

\begin{figure}[t]
    \centering
    \includegraphics[width=0.99\columnwidth]{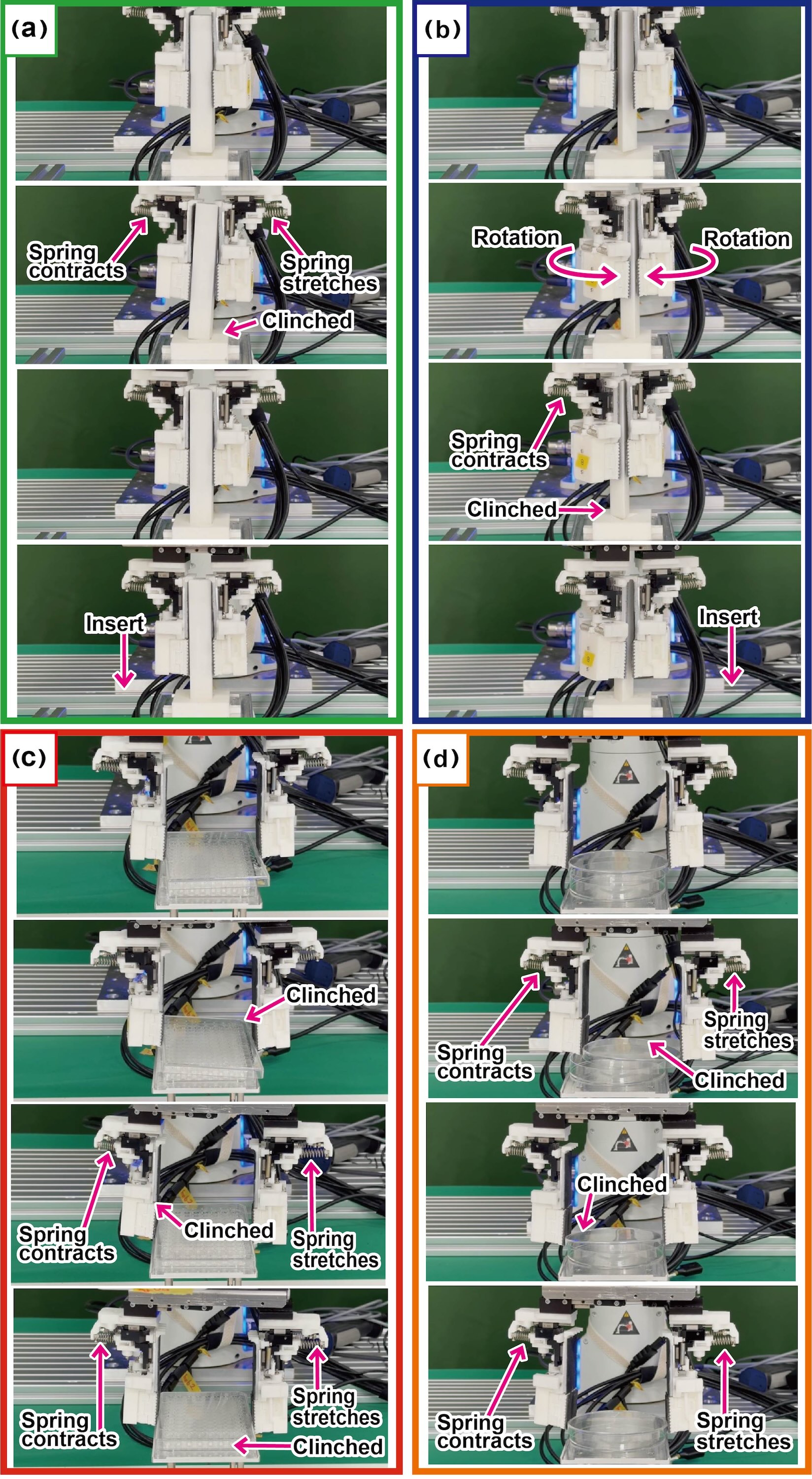}
    \caption{
    Insertion process for (a) square prisms, (b) triangular prisms, (c) well-plates, and (d) petri dishes. 
    }
    \label{fig:result_peg_in_test}
\end{figure}

Offsets of $\pm$4~mm in the $x_T$ and $y_T$ directions were implemented with referencing to the existing literature~\cite{Zhang_6DOF_spral_PEG_insert}. Although this reference literature does not add any perturbation to the yaw axis, a tilt was added in our experiments around the $yaw_T$-axis to confirm the effect of the proposed mechanism. This tilt was set at $\pm8$ degrees, which exceeds the insertion limit angle established in \cref{sec:result_insert_angle_Yaw}, and eight different experiments were conducted combining these factors ($\pm{x_t},\pm{y_t},\pm{yaw_t}$).
The spiral path necessitates that the grasped object passes above the insertion site at some point during its course. 
However, in reality, because the adaptive mechanism responds to friction or snagging between the grasping object and the contact surface, the object held by the robot hand did not always traverse above the insertion site, even with a spiral curve radius similar in size to cover the offset amount.
For this reason, the final radius value took into account not only the offset amounts but also the adaptive variation range (maximum displacement of 13~mm in the $y$-axis direction). The set conditions included a starting radius 5~mm, end radius of 18~mm, a circumference of 12 laps, and a pushing force of 11~N.
To validate how the adaptive function impacts the results, we also conducted a comparative experiment with the $x$-axis and yaw-axis adaptive functions disabled.
The results are tabulated in Table~\ref{tab:exp_result_square_prism_insert}.

\begin{table}[t]
    \centering
    \caption{Results of peg-in-hole insertion using a square prism with adaptive functions enabled/disabled.}
    \label{tab:exp_result_square_prism_insert}
    \begingroup
    \scalefont{0.85}
    \begin{tabular}{c||cccccccc}
    \hline
    \multicolumn{1}{c||}{Adaptive axis} &
    \multicolumn{8}{c}{Positive/negative offset of $x_T, y_T,yaw_T$}\\
    \cline{2-9}
    \multicolumn{1}{c||}{x~~y~~z~~yaw} &
          +++ & +-+ & -++ & - - + & + + - & + - - & - + - & - - - \\
        \hline
        \hline
        ~$\surd$~$\surd$~$\surd$~$\surd$~ & \bf{5/5} & \bf{5/5} & \bf{5/5} & \bf{5/5} & \bf{5/5} & \bf{5/5} & \bf{5/5} & \bf{5/5} \\ 
        ~---~$\surd$~$\surd$~$\surd$~ & \bf{5/5} & \bf{5/5} & \bf{5/5} & \bf{5/5} & \bf{5/5} & \bf{5/5} & 2/5 & 0/5 \\ 
        ~$\surd$~$\surd$~$\surd$~---~ & \bf{5/5} & \bf{5/5} & \bf{5/5} & \bf{5/5} & 0/5 & 0/5 & \bf{5/5} & 0/5 \\ 
        ~---~$\surd$~$\surd$~---~ & \bf{5/5} & \bf{5/5} & \bf{5/5} & \bf{5/5} & 0/5 & 0/5 & 0/5 & 0/5 \\ 
        ~$\surd$~---~$\surd$~$\surd$~ & 0/5 & 0/5 & 2/5 & 0/5 & 0/5 & 0/5 & 0/5 & 0/5 \\ 
        ~$\surd$~$\surd$~---~$\surd$~ & \bf{5/5} & 4/5 & 0/5 & \bf{5/5} & \bf{5/5} & \bf{5/5} & \bf{5/5} & 4/5 \\ 
        ~---~---~---~---~         & 0/5 & 0/5 & 0/5 & 0/5 & 0/5 & 0/5 & 0/5 & 0/5 \\ 

    \hline
    \end{tabular}
    \endgroup
\end{table}

These results indicate that with all adaptive functions engaged, the success rate is 100$\%$, which compares to 62.5$\%$ when the yaw-axis is locked, 50$\%$ with $x$ and yaw-axes locked, and 80$\%$ when the $x$-axis alone is locked. Failures predominantly occurred when the yaw-axis error pointed in the negative direction. 
Additionally, we note that all failures occurred with a negative offset to the yaw-axis.

This is attributed to the fact that when the $yaw_T$-axis of offset is positive, the inclines of the spiral path and the square prism follow the same direction, resulting in a relatively minor hole misalignment. However, this misalignment doubles in the negative direction. 
As a result, failures are more likely when the adaptive around the yaw-axis is disabled. Even if the orientation around the yaw-axis of the finger aligns with the hole, insertion cannot occur unless all four sides of the quadrilateral are parallel to the hole. The failure observed during the $x$-axis lock occurred because, even if the yaw-axis was functioning, it couldn't adequately handle these large misalignment errors unless the $x$-axis adaptive was also operational.
The success rate when the $z$-axis was locked was moderately high at 82.5$\%$. Failures occurred primarily because, despite the pressing force being equalized by the F/T sensor, the square prism occasionally detached from the hole due to the inability to absorb the $z$-axis impact when catching in the hole. Furthermore, in a few instances, the operation halted because the load in the $z_{H}$ direction exceeded the hand's limit during operation, a result of the robot hand's disabled $z$-axis adaptive function, which left it unable to absorb the initial contact shock.
Nearly all attempts failed when the $y$-axis was locked. With the support of the yaw axis adaptive, the robot hand could insert it in only twice.
When all four axes were locked, the success rate plummeted to 0$\%$.

\subsubsection{Insertion Experiment of Triangular Prism}
\label{sec:result_triangle_prism}

\begin{table}[t]
    \centering
    \caption{Results of peg-in-hole insertion using a triangle prism with adaptive functions enabled/disabled.}
    \label{tab:exp_result_triangle_prism_insert}
    \begingroup
    \scalefont{0.85}
    \begin{tabular}{c||cccccccc}
    \hline
    \multicolumn{1}{c||}{Adaptive axis} &
    \multicolumn{8}{c}{Positive/negative offset of $x_T, y_T,yaw_T$}\\
    \cline{2-9}
    \multicolumn{1}{c||}{x~~y~~z~~yaw} &
          +++ & +-+ & -++ & - - + & + + - & + - - & - + - & - - - \\
         \hline
        \hline
        ~$\surd$~$\surd$~$\surd$~$\surd$~ & \bf{5/5} & \bf{5/5} & 3/5 & \bf{5/5} & \bf{5/5} & 4/5 & 3/5 & 5/5 \\ 
        ~$\surd$~$\surd$~$\surd$~---~ & 2/5 & 3/5 & 1/5 & \bf{5/5} & \bf{5/5} & 2/5 & \bf{5/5} & 3/5 \\ 
        ~---~$\surd$~$\surd$~$\surd$~ & 1/5 & \bf{5/5} & 4/5 & 0/5 & \bf{5/5} & 4/5 & 0/5 & 2/5 \\ 
        ~---~$\surd$~$\surd$~---~ & 3/5 & 0/5 & 1/5 & 4/5 & \bf{5/5} & 4/5 & \bf{5/5} & 4/5 \\ 
    \hline
    \end{tabular}
    \endgroup
\end{table}

In order to add to the stability during grasping, the pegs were grasped in the posture that employed the yaw axis rotation function depicted in Fig.~\ref{fig:adaptive_motion}~(f).
The offset was established at $\pm$ 4~mm in the $x_T$ and $y_T$ directions in alignment with the reference literature~\cite{Zhang_6DOF_spral_PEG_insert}, while a perturbation of $\pm15$ degrees was set on the $yaw_T$-axis. 
A triangular prism could not be inserted in the same spiral path as a square prism due to the offset value of yaw-axis of a triangular prism being 1.875 times that of a square prism, compounded by the challenging nature of its shape for insertion.
Hence, the final radius was expanded to 26~mm, which is nearly 1.45 times larger than that of the square prism.
The set conditions included a starting radius of 7~mm, an ending radius of 26~mm, a spiral encompassing 8 laps, and a pushing force of 12~N (Fig.~\ref{fig:result_peg_in_test}~(b)).
We altered the posture of the triangular prism while grasping it, clutching the prism in the grip configuration illustrated in Fig.\ref{fig:x_Yaw_function}~(b). 
In terms of the impact of immobilizing each adaptive mechanism, the results from the square prisms insertion experiment suggested a detrimental effect when the $z$ and $y$ axes were locked as shown in \cref{sec:result_square_prism}. Consequently, we did not investigate the impact of locking the adaptive function for the $y$-axis and $z$-axis.
The results are tabulated in Table~\ref{tab:exp_result_triangle_prism_insert}.
The experimental results revealed that the success rate using all adaptive features was 87.5$\%$. 

When activated, the yaw axis adaptive mechanism increases the contact area with the triangular prism due to fingertip tilt as shown in Fig.~\ref{fig:x_Yaw_function}~(b). The functioning of the $x$-axis allows dynamic movement along the side of the prism. Subsequently, the FAAF Hand can adjust the prism's posture within the finger upon application of external forces. However, significant posture changes arise if the prism catches on the edge from the back, making the insertion problematic, which resulted in a success rate of 87.5$\%$ despite all mechanisms operating.
Particularly when the $x$-axis is fixed, the influence on posture changes upon contact with the edge intensifies, leading to a reduced success rate of 52.5$\%$. With a yaw-axis fix, the $x$-axis performs a pseudo-yaw function and supports the prism’s posture alteration (Fig.~\ref{fig:x_Yaw_function}~(c)), resulting in a success rate of 77.5$\%$.
When both the $x$- and yaw-axes are fixed, grasping occurs as shown in Fig.~\ref{fig:x_Yaw_function}~(a). The activation of the $y$-axis's adaptive mechanism enables some level of posture readjustment, yielding a comparable success rate of 77.5$\%$.
Posture instability due to impacts is more pronounced for the triangular prism than the square prism. If an increased success rate is desired, potential improvements include refining fingertip protrusions to extend in both horizontal and vertical directions. However, since these protrusions might interfere during the grasp of other objects, their inclusion should be decided based on intent.

\subsubsection{Lateral Insertion}
\label{sec:result_lateral_insert}

\begin{figure}[t]
    \centering
    \includegraphics[width=0.95\columnwidth]{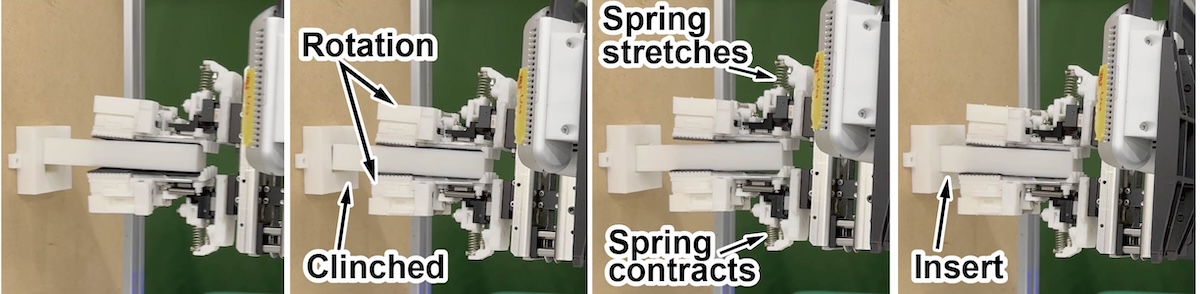}
    \caption{
    Insertion process for the square prism from the lateral grasp position.
    }
    \label{fig:result_side_insert}
    \vspace{2mm}
\end{figure}

\begin{table}[t]
    \centering
    \caption{Experimental results of the square prism at lateral grasp.}
    \label{tab:exp_result_lateral_grasp_triangle_prism_insert}
    \begingroup
    \scalefont{0.85}
    \begin{tabular}{c||cccccccc}
    \hline
    \multicolumn{1}{c||}{Adaptive axis} &
    \multicolumn{8}{c}{Positive/negative offset of $x_T,y_T,yaw_T$}\\
    \cline{2-9}
    \multicolumn{1}{c||}{x~~y~~z~~yaw} &
          +++ & +-+ & -++ & - - + & + + - & + - - & - + - & - - - \\
         \hline
        \hline
        ~$\surd$~$\surd$~$\surd$~$\surd$~ & \bf{5/5} & \bf{5/5} & \bf{5/5} & \bf{5/5} & \bf{5/5} & \bf{5/5} & \bf{5/5} & \bf{5/5} \\  
        ~$\surd$~$\surd$~$\surd$~---~ & 0/5 & 0/5 & 0/5 & \bf{5/5} & \bf{5/5} & \bf{5/5} & \bf{5/5} & \bf{5/5} \\ 
        ~---~$\surd$~$\surd$~$\surd$~  & 1/5 & \bf{5/5} & \bf{5/5} & 2/5 & \bf{5/5} & \bf{5/5} & \bf{5/5} & \bf{5/5} \\ 
        ~---~$\surd$~$\surd$~---~ & 0/5 & 0/5 & 0/5 & 0/5 & \bf{5/5} & \bf{5/5} & \bf{5/5} & \bf{5/5} \\ 
    \hline
    \end{tabular}
    \endgroup
\end{table}

Given that the adaptive element is solely concentrated within the lightweight finger segment, lateral insertion can be achieved in a similar fashion to vertical insertion, as presented in Fig.~\ref{fig:result_side_insert}.
In the results of~\ref{sec:result_square_prism}, it was observed that $yaw_T$ is prone to failure when the direction of tilt of $yaw_T$ is opposite to the direction of rotation of the spiral path. 
Therefore, we verified whether the failure condition also reverses when the direction of rotation of the spiral path is reversed. 
The results are presented in Table~\ref{tab:exp_result_lateral_grasp_triangle_prism_insert}, with the trend closely mirroring that displayed in Table~\ref{tab:exp_result_square_prism_insert}. Failures are concentrated when $yaw_T$ is positive because the direction of rotation of the spiral path is opposite to that of \cref{sec:result_square_prism}.
This confirms that the experimental method can be used to insert pegs in postures other than top-grasp.


\subsection{Lid Insertion}
\label{sec:result_insert_well_plate_lid}

\begin{table}[t]
    \centering
    \caption{Results of lid insertion for well-plate and petri dish with all adaptive functions enabled.}
    \label{tab:exp_result_normal_well_plate}
    \begingroup
    \scalefont{0.8}
    \begin{tabular}{c||cccccccc}
    \hline
    \multicolumn{1}{c||}{Offset} & {}\\
    \multicolumn{1}{c||}{($x_T,y_T,yaw_T$)} &  \multicolumn{8}{c}{Positive/negative offset of $x_T, y_T,yaw_T$}\\
    \cline{2-9}
    \multicolumn{1}{c||}{([mm], [mm], [deg])} &
         +++&+-+&-++&- -+&++-&+- -&-+-&- - -\\
        \hline
        \hline
        \multicolumn{1}{c||}{Well-plate lid} &{}\\
        $(\pm2, \pm2, \pm2)$ & \bf{5/5} & \bf{5/5} & \bf{5/5} & \bf{5/5} & \bf{5/5} & \bf{5/5} & \bf{5/5} & \bf{5/5} \\
        $(\pm4, \pm4, \pm4)$ & \bf{5/5} & \bf{5/5} & \bf{5/5} & \bf{5/5} & \bf{5/5} & \bf{5/5} & \bf{5/5} & \bf{5/5} \\
        $(\pm6, \pm6, \pm6)$ & \bf{5/5} & \bf{5/5} & \bf{5/5} & \bf{5/5} & \bf{5/5} & \bf{5/5} & \bf{5/5} & \bf{5/5} \\
        \hline

        \multicolumn{1}{c||}{Petri dish lid} &{}\\
         $(\pm6, \pm6, \pm6)$ & \bf{5/5} & \bf{5/5} & \bf{5/5} & \bf{5/5} & \bf{5/5} & \bf{5/5} & \bf{5/5} & \bf{5/5} \\
        
    \hline
    \end{tabular}
    \endgroup
\end{table}

The degree of offset was established at 6~mm in the $x_T$ and $y_T$ directions, an increase of 2~mm compared to the offset in the peg experiment to account for the 2~mm clearance between the lid and the well-plate. For the yaw axis rotation ($yaw_T$), 6 degrees were selected, corresponding to twice the insertion limit of 3 degrees established in \cref{sec:exp_set_insert_angle_Yaw}.
The experimental results validated that successful insertion could be conducted for all eight combinations of spiral path, utilizing a starting radius of 7~mm, an ending radius of 16~mm, a rotation of 8 laps, and a consistent pushing force of 8N. Despite presenting a greater positional error than the peg, the termination radius is not correspondingly large. This is due to the difference in insertion methods between Peg and Lid.
To confirm whether insertion is possible even with different offset amounts, the same experiment was conducted with offset amounts of $\pm4$ and $\pm2$~mm in the $x_T$ and $y_T$ directions and offset amounts of 4 degrees and 2 degrees around the yaw-axis ($yaw_T$), and all of them could be inserted in the same Table~\ref{tab:exp_result_normal_well_plate}.
When the lid is forced downwards along the spiral path, the interior angle of the lid catches on the step due to the automatic posture change around the roll axis, as illustrated in Fig.~\ref{fig:adaptive_grasp_work}~(g).
Subsequently, as the spiral path persists, each adaptive mechanism continues to retain this trapped angle through its operation, as featured in Fig.~\ref{fig:adaptive_grasp_work}~(b)-(e). This snagged corner subsequently dictates the movement of the lid, enhancing the chances of hooking onto other corners. As a result, the second and third corners sequentially become trapped, culminating in all corners becoming stuck (Fig.~\ref{fig:result_peg_in_test}~(c)). 
Insertion experiments were also performed for petri dish lids under the same conditions.
In the case of the circular shape, the effect of the tilt around the $yaw_T$ axis was smaller, and the circular shape could be inserted more smoothly than the well-plate lid because the edges of the circular shape were easier to catch than the square shape.

\begin{table}[t]
    \centering
    \caption{Results of well-plate lid insertion with adaptive functions enabled/disabled.}
    \label{tab:exp_result_adapt_locked_well_plate}
    \begingroup
    \scalefont{0.85}
    \begin{tabular}{c||cccccccc}
    \hline
    \multicolumn{1}{c||}{Adaptive axis} &
    \multicolumn{8}{c}{Positive/negative offset of $x_T, y_T, yaw_T$}\\
    \cline{2-9}
    \multicolumn{1}{c||}{x~~y~~z~~yaw} &
          +++ & +-+ & -++ & - - + & + + - & + - - & - + - & - - - \\
         \hline
        \hline
    ~$\surd$~~$\surd$~~$\surd$~~$\surd$~ & \bf{5/5} & \bf{5/5} & \bf{5/5} & \bf{5/5} & \bf{5/5} & \bf{5/5} & \bf{5/5} & \bf{5/5} \\ 
    ~$\surd$~---~$\surd$~$\surd$~ & 0/5 & \bf{5/5} & \bf{5/5} & 3/5 & 1/5 & 0/5 & 0/5 & 0/5 \\ 
    ~$\surd$~~$\surd$~---~$\surd$~ & 3/5 & \bf{5/5} & \bf{5/5} & \bf{5/5} & \bf{5/5} & 3/5 & \bf{5/5} & \bf{5/5} \\ 
    ~---~$\surd$~$\surd$~~$\surd$~ & \bf{5/5} & \bf{5/5} & \bf{5/5} & \bf{5/5} & \bf{5/5} & \bf{5/5} & \bf{5/5} & \bf{5/5} \\  
    ~$\surd$~$\surd$~$\surd$~---~ & \bf{5/5} & \bf{5/5} & \bf{5/5} & \bf{5/5} & \bf{5/5} & \bf{5/5} & \bf{5/5} & \bf{5/5} \\ 
    ~---~$\surd$~$\surd$~--- & \bf{5/5} & \bf{5/5} & \bf{5/5} & \bf{5/5} & \bf{5/5} & \bf{5/5} & \bf{5/5} & \bf{5/5} \\ 
    ~---~---~$\surd$~---~ & 0/5 & 3/5 & 0/5 & 0/5 & 0/5 & 0/5 & 0/5 & 0/5 \\ 
    \hline
    \end{tabular}
    \endgroup
\end{table}
    
Insertion experiments were performed under the same conditions with each adaptive mechanism of the finger fixed (amount of offset: $x, y$ axis 6~mm, 6 degrees around yaw axis).
When we performed experiments with the $x$-axis and yaw-axis fixed, all insertions were successful. This is due to the assistance provided by the $y$-axis adaptive mechanism to the necessary yaw-axis direction rotation and a fingertip slippage takes the place of the lateral adaptive function of the lid for insertion. However, these results do not rule out the possibility that all insertion motions can be executed without the finger adaptive mechanism by moving the hand along a spiral path.
To confirm the effectiveness of the adaptive function, additional experiments were performed with the $y$-axis and $z$-axis locked. Results showed a high failure rate under many conditions with the $y$-axis locked. A particularly pronounced tendency to fail was observed when the $yaw{_T}$ was negative, reminiscent of square prism insertions.
Failures were also recorded when the $z$-axis was locked. Despite the $z$-axis moving in a manner akin to adaptivity by maintaining constant pressing force via the F/T sensor, it fails to provide the tilt in the roll direction as depicted in  Fig.~\ref{fig:adaptive_grasp_work}~(g).
Therefore, it was determined that the y and $z$-axis adaptive functions are important for lid insertion.


\section{CONCLUSIONS}
This paper explores the following enhancements to expand the range of use of parallel gripper-type hands operated by a single actuator:

\begin{itemize}

   \item Incorporate multiple passive mechanisms exclusively in the lightweight finger section.
    \item Develop mechanisms that allow passive in-hand manipulation when grasped, in response to external forces.
    \item Passivity is provided to facilitate object insertion, even when there is a positional error in all directions and rotational error along the $y$-axis, between the target site and the object.

\end{itemize}

To achieve these, we designed the FAAF Hand, equipped with an adaptive finger structure. 
To confirm the usefulness of these mechanisms, we conducted an insertion experiment employing a simple control method: drawing an arbitrary spiral path with the robot hand.
The experiment proved successful in inserting a square or triangular prism, demonstrating automatic posture adjustment to align with the hole despite displacements in 2 directions ($x$=4~mm, $y$=4~mm) parallel to the table, and the postural error along the central axis (yaw=8 and 15 deg).
Furthermore, the robot hand's adaptive response being confined to the lightweight finger unit allows for versatile insertion postures, including lateral operation.
The well-plate lids and petri dish lid, offset by 6~mm and 6 degrees around the yaw-axis, were also successfully inserted employing the same simple control method.


\bibliographystyle{IEEEtran} 
\bibliography{FAAF_hand_av}

\end{document}